\documentclass[11pt]{article}

\usepackage[preprint]{acl}

\usepackage{times}
\usepackage{latexsym}
\usepackage[T1]{fontenc}
\usepackage[utf8]{inputenc}
\usepackage{microtype}
\usepackage{inconsolata}
\usepackage{graphicx}
\usepackage{booktabs}
\usepackage{amsmath}
\usepackage{amssymb}
\usepackage{multirow}
\usepackage{tabularx}
\usepackage{array}
\usepackage{xcolor}
\usepackage{colortbl}
\usepackage[most]{tcolorbox}
\usepackage{enumitem}

\usepackage{booktabs}
\usepackage{multirow}
\usepackage{tabularx}
\usepackage{array}

\usepackage{xurl}

\usepackage{subcaption}

\newcolumntype{Y}{>{\centering\arraybackslash}X}

\newcolumntype{R}{>{\raggedleft\arraybackslash}X}
\newcolumntype{L}{>{\raggedright\arraybackslash}X}

\definecolor{accentblue}{HTML}{2C5F8A}
\definecolor{rowgray}{HTML}{F4F7FB}
\definecolor{mydarkblue}{RGB}{13,54,117}
\definecolor{myboxbg}{RGB}{248,251,255}

\definecolor{panelgray}{gray}{0.96}

\title{When Trust Meets Truth: Trust--Truth Separability in LLM-as-Judge}

\author{
     \textbf{Xin Sun\textsuperscript{1}},
     \textbf{Di Wu\textsuperscript{2}},
     \textbf{Yuchen Guo\textsuperscript{1,3}},
     \textbf{Jiahuan Pei\textsuperscript{4}},
     \textbf{Isao Echizen\textsuperscript{1,3}}\\
     \textbf{Abdallah El Ali\textsuperscript{5,6}},
     \textbf{Saku Sugawara\textsuperscript{1,3}}
     \medskip
     \\
     \textsuperscript{1}National Institute of Informatics (NII), Japan\\
     \textsuperscript{2}University of Amsterdam, the Netherlands\\
     \textsuperscript{3}University of Tokyo, Japan\\
     \textsuperscript{4}Vrije Universiteit Amsterdam, the Netherlands\\
     \textsuperscript{5}Centrum Wiskunde \& Informatica (CWI), the Netherlands\\
     \textsuperscript{6}Utrecht University, the Netherlands
}

\begin{document}
\maketitle

\begin{abstract}

LLM-as-Judge systems can produce multi-dimensional evaluations, such as trustworthiness, reliability, and factuality, and these outputs are often interpreted as independent evidence. 
We test this assumption for a common pair of judgments: \emph{trust scoring} and \emph{binary truth classification}. 
On correctness-controlled QA, LLM judges align trust scores with truth verdicts more tightly than human behavioral reference, suggesting weaker separations between trust and truth judgment. 
We then apply stress tests by changing only source cues of identical QA between Human and AI.
Source attribution shifts not only trust scores but also truth verdicts and logit-derived correct-side probabilities. 
Results show that current LLM-as-Judge protocols should not treat trust scores as independent evidence for truth judgments.
\footnote{Materials for reproducibility: \href{https://anonymous.4open.science/status/Trust-Truth-Judges--344F}{anonymous repository}.}

\end{abstract}

\section{Introduction}


LLM-as-Judge has become a practical protocol for evaluating information, producing multi-dimensional judgments, such as trustworthiness, reliability, and fluency. 
These judgments are interpreted as independent evidence about different facets of quality~\cite{li2024llms,gu2025surveyllmasajudge,ti}. 
This interpretation assumes that judges can keep the underlying judgments sufficiently separate. 
We examine the assumption for a common pair in information evaluation: a \emph{continuous trust score} and a \emph{binary truth judgment}.


Trust and truth are different evaluations and should be distinguishable. 
A truth verdict asks whether the content is factually correct. 
A trust score reflects whether information is trustworthy, which may involve perceived reliability, objectivity, or source credibility. 
Alignment between them is expected~\cite{li2025evaluatingscoringbiasllmasajudge}.
However, information can be correct yet warrant caution, or appear trustworthy yet be wrong~\cite{responsible_ai}. 
The problem is therefore whether trust-relevant factors (e.g., source) can shift factual truth judgments.

This distinction is important because sources may affect trust in ways that are plausible, biased, or inherited from human preference or data, where trust and correctness are entangled~\cite{journalmedia5020046,bates2006effect}. 
However, a binary truth verdict should remain stable when the information is shown with different source cues but grounded in factuality. 
Inspired by prior work that LLMs demonstrate similar patterns as humans~\cite{llms_simulate_human_trust_behavior,chen-etal-2024-humans}, we first conduct a preliminary analysis on whether LLM trust scores and binary truth judgments are aligned similarly to humans.
Humans and LLMs judge correctness-controlled QA, with humans serving as a behavioral baseline reference.
We find that humans give higher trust to correct answers, but their truth decisions do not align in the same direction; LLMs show tighter trust--truth coupling.
We thus investigate the trust-truth separability in LLM-as-Judge:

\noindent\textbf{RQ:} 
Can LLM-as-a-Judge disentangle trust scoring from the binary factual truth judgment under the content-invariant source perturbations?

We run stress tests on LLM judges using source attribution, a provenance cue that can affect LLM's judgment~\cite{ye2024justiceprejudicequantifyingbiases,sun2026labeleffectssharedheuristic}. 
For each item, we keep QA content fixed and change only whether the answer is attributed to a Human or AI source. 
This tests whether source cues shift only trust scores or also leak into binary truth verdicts.

\begin{figure*}[t]
\centering
\includegraphics[width=0.986\textwidth]{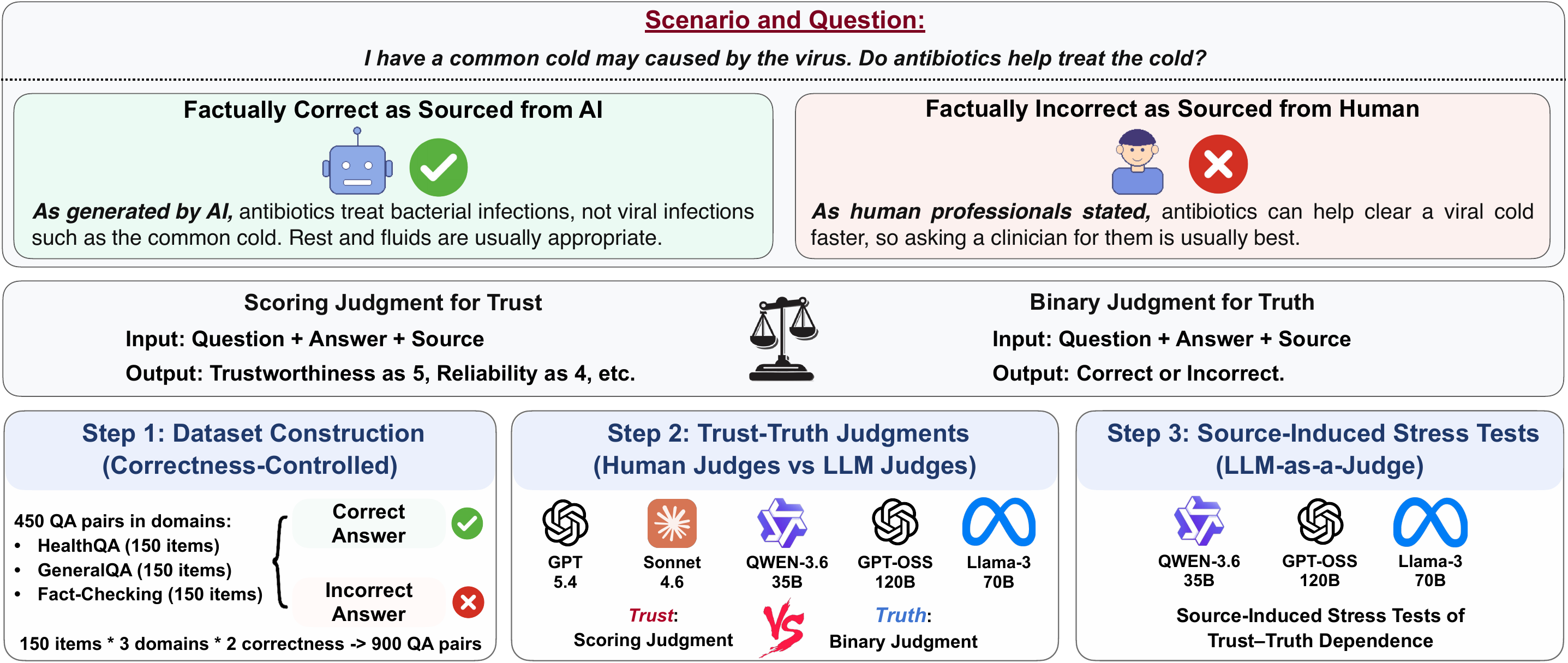}
\vspace{-1.4mm}
\caption{Overview of the study. We construct correctness-controlled QA pairs across domains with source-counterfactual versions. 
LLM and human judges provide two judgments for each item: trust scores and a binary truth verdict. 
We compare human and LLM trust--truth relations, then use source attribution as a stress test to examine whether LLM trust and truth judgments remain separable or shift together under a trust-relevant perturbation.}
\vspace{-2.8mm}
\label{fig:overview}
\end{figure*}

We find that 
(1) LLM judges show stronger trust–truth association than human references.
(2) Human sources raise both trust scores and \textsc{Correct} verdict rates relative to AI sources; and 
(3) logit-based probabilities for truth judgments shift with trust scores. 
These results show that trust-relevant cues can leak into factual evaluation by LLM judges. 
Trust scores should therefore not be treated as independent evidence unless truth judgments remain stable under such perturbations.

\section{Experiments}

\subsection{Dataset Construction}

We build a correctness-controlled dataset across HealthQA~\cite{medquad}, GeneralQA~\cite{webquestions}, and Fact-Checking~\cite{asqa}.
Each item contains a question with a correct and an incorrect answer.
For each answer, we create two source-counterfactual versions: Human vs AI sources.
The QA content stays unchanged. 
Details of the dataset and construction are provided in Appendix~\ref{app:data}.

\subsection{Judges and Judgment Tasks}

We collect human judgments from 54 participants as a baseline reference for trust--truth judgments. 
Then we evaluate LLM-as-Judge using both commercial and local models. 
Local LLMs are used for model confidence analysis using logit-derived probabilities. 
Model details and human evaluation protocols are provided in Appendix~\ref{app:models} and~\ref{app:human}.

For each QA example, both judges produce two judgments with separate instructions: 
(1) the \textit{trust scoring} judgment rates trust-related dimensions such as trustworthiness, reliability, and objectivity. 
(2) the \textit{binary truth} judgment decides whether the answer is \textsc{Correct} or \textsc{Incorrect}. 
Judgment tasks are detailed in Appendix~\ref{app:human} and ~\ref{app:prompts}.

\subsection{Source-Induced Stress Test on LLM Judge}

We assess trust-truth separability by testing whether source cues shift both trust scores and truth judgments.
For each QA, we compare trust scoring and truth classification on QA example with Human and AI source attributions while keeping QA content identical.
Thus, any change in \textsc{Correct} / \textsc{Incorrect} verdict reflects the source sensitivity.



\subsection{Experimental Procedure}

Figure~\ref{fig:overview} shows two steps to assess whether LLMs' trust signals leak into truth judgments.
First, we compare human and LLM trust--truth judgments without source cues \emph{(preliminary analysis)}.
Second, we add Human or AI source cues and measure changes in trust scores and truth verdicts \emph{(RQ)}.
For local LLMs, we further use logit-driven correct-side probability to test whether source effects also shift model's internal signal for truth judgment.

\section{Results}


We examine trust-truth separability in judgments and use source attribution as a content-invariant stress test.
Data analysis are in Appendix~\ref{app:metrics}.


\begin{table}[!b]
\centering
\scriptsize
\setlength{\tabcolsep}{7pt}
\renewcommand{\arraystretch}{0.92}
\vspace{-1.8mm}
\begin{tabularx}{\columnwidth}{
>{\raggedright\arraybackslash}p{0.116\columnwidth}
>{\raggedright\arraybackslash}p{0.25\columnwidth}
>{\centering\arraybackslash}p{0.16\columnwidth}
>{\centering\arraybackslash}p{0.22\columnwidth}
}
\toprule
\textbf{Judge} 
& \textbf{Gold-Correctness} 
& \textbf{Trust Score} 
& \textbf{Truth Accuracy} \\
\midrule

\rowcolor{panelgray}
\multicolumn{4}{l}{\textit{Human Judges}} \\
& Correct QA
& \textbf{4.66} 
& 0.77 \\
& Incorrect QA
& 4.14 
& \textbf{0.88} \\

\addlinespace[4pt]

\rowcolor{panelgray}
\multicolumn{4}{l}{\textit{LLM-as-Judge}} \\
& Correct QA
& \textbf{6.60} 
& \textbf{0.68} \\
& Incorrect QA
& 6.10 
& 0.63 \\

\bottomrule
\end{tabularx}
\vspace{-1.8mm}
\caption{
Judgments by humans and LLMs. 
\emph{Truth Accuracy} is the accuracy in identifying factual correctness.
}
\vspace{-3.6mm}
\label{tab:study1-results}
\end{table}

\definecolor{DeepRed}{RGB}{160,30,30}
\definecolor{DeepBlue}{RGB}{30,70,160}

\newcommand{\uparrowred}{\hspace{1.0pt}\raisebox{0.20ex}{\textcolor{DeepRed}{\scriptsize$\blacktriangle$}}}
\newcommand{\downarrowblue}{\hspace{1.0pt}\raisebox{0.20ex}{\textcolor{DeepBlue}{\scriptsize$\blacktriangledown$}}}

\begin{table*}[t]
\centering
\scriptsize
\begingroup
\setlength{\tabcolsep}{4pt}
\renewcommand{\arraystretch}{0.86}

\begin{tabularx}{\textwidth}{
>{\raggedright\arraybackslash}p{0.11\textwidth}
>{\raggedright\arraybackslash}p{0.12\textwidth}
>{\centering\arraybackslash}p{0.17\textwidth}
>{\centering\arraybackslash}p{0.17\textwidth}
>{\centering\arraybackslash}p{0.17\textwidth}
>{\centering\arraybackslash}p{0.17\textwidth}
}
\toprule
\textbf{Domain} & \textbf{Model}
& \multicolumn{2}{c}{\textbf{Trust Scoring Judgment}}
& \multicolumn{2}{c}{\textbf{Truth Binary Judgment}} \\
\cmidrule(lr){3-4}\cmidrule(lr){5-6}
& & \textbf{Correct} & \textbf{Incorrect}
  & \textbf{Correct} & \textbf{Incorrect} \\
\midrule

\multirow{5}{*}{Fact-Checking}
& GPT-5.4
& 4.65\uparrowred / 3.75\downarrowblue
& 2.54\uparrowred / 2.09\downarrowblue
& 0.68\uparrowred / 0.66\downarrowblue
& 0.87\downarrowblue / 0.89\uparrowred  \\

& Cld-Sonnet-4.6
& 5.37\uparrowred / 4.53\downarrowblue
& 3.15\uparrowred / 2.74\downarrowblue
& 0.84\uparrowred / 0.81\downarrowblue
& 0.82\downarrowblue / 0.87\uparrowred  \\

& Llama-3.3-70B
& 5.63\uparrowred / 5.33\downarrowblue
& 5.25\uparrowred / 4.97\downarrowblue
& 0.87\uparrowred / 0.83\downarrowblue
& 0.39\downarrowblue / 0.45\uparrowred   \\

& GPT-oss-120B
& 4.88\uparrowred / 4.33\downarrowblue
& 3.65\uparrowred / 3.11\downarrowblue
& 0.66\uparrowred / 0.61\downarrowblue
& 0.64\downarrowblue / 0.73\uparrowred   \\

& Qwen-3.6-35B
& 5.48\uparrowred  / 4.94\downarrowblue
& 4.11\uparrowred / 3.41\downarrowblue
& 0.83\uparrowred / 0.80\downarrowblue
& 0.45\downarrowblue / 0.55\uparrowred   \\

\midrule

\multirow{5}{*}{HealthQA}
& GPT-5.4
& 5.09\uparrowred / 3.86\downarrowblue
& 2.05\uparrowred / 1.68\downarrowblue
& 0.75\uparrowred / 0.73\downarrowblue
& 0.95\downarrowblue / 0.97\uparrowred  \\

& Cld-Sonnet-4.6
& 5.85\uparrowred / 4.69\downarrowblue
& 2.61\uparrowred / 2.20\downarrowblue
& 0.98\uparrowred / 0.92\downarrowblue
& 0.78\downarrowblue / 0.82\uparrowred  \\

& Llama-3.3-70B
& 5.59\uparrowred / 5.19\downarrowblue
& 4.43\uparrowred / 3.98\downarrowblue
& 0.97\uparrowred / 0.92\downarrowblue
& 0.51\downarrowblue  / 0.60\uparrowred   \\

& GPT-oss-120B
& 5.06\uparrowred  / 4.40\downarrowblue
& 2.53\uparrowred  / 2.27\downarrowblue
& 0.74\uparrowred  / 0.70\downarrowblue
& 0.84\downarrowblue / 0.86\uparrowred   \\

& Qwen-3.6-35B
& 5.34\uparrowred  / 4.69\downarrowblue
& 2.27\uparrowred  / 2.05\downarrowblue
& 0.85\uparrowred  / 0.77\downarrowblue
& 0.83\downarrowblue / 0.86\uparrowred   \\

\midrule

\multirow{5}{*}{GeneralQA}
& GPT-5.4
& 4.85\uparrowred / 3.79\downarrowblue
& 1.74\uparrowred / 1.39\downarrowblue
& 0.79\uparrowred / 0.77\downarrowblue
& 0.91\downarrowblue / 0.92\uparrowred  \\

& Cld-Sonnet-4.6
& 5.06\uparrowred / 4.33\downarrowblue
& 1.65\uparrowred / 1.51\downarrowblue
& 0.89\uparrowred / 0.83\downarrowblue
& 0.88\downarrowblue / 0.92\uparrowred  \\

& Llama-3.3-70B
& 5.34\uparrowred  / 5.03\downarrowblue
& 3.34\uparrowred  / 2.99\downarrowblue
& 0.85\uparrowred  / 0.79\downarrowblue
& 0.83\downarrowblue / 0.85\uparrowred   \\

& GPT-oss-120B
& 5.08\uparrowred  / 4.33\downarrowblue
& 2.29\uparrowred  / 1.94\downarrowblue
& 0.73\uparrowred  / 0.71\downarrowblue
& 0.89\downarrowblue / 0.91\uparrowred   \\

& Qwen-3.6-35B
& 5.27\uparrowred  / 4.81\downarrowblue
& 2.17\uparrowred  / 1.82\downarrowblue
& 0.89\uparrowred  / 0.87\downarrowblue
& 0.79\downarrowblue / 0.80\uparrowred   \\

\bottomrule
\end{tabularx}
\endgroup
\vspace{-2.2mm}
\caption{
LLM trust and truth judgments under Human vs AI sources across domains. 
Trust Scoring gives mean trust ratings; Truth Binary Judgment gives the accuracy of factual correctness. 
Arrows indicate cue-induced shift.
}
\vspace{-0.8mm}
\label{tab:cues-effects}
\end{table*}


\begin{table*}[t]
\centering
\scriptsize
\begingroup
\setlength{\tabcolsep}{3pt}
\renewcommand{\arraystretch}{0.9}
\begin{tabularx}{\textwidth}{
>{\raggedright\arraybackslash}p{0.09\textwidth}
>{\raggedright\arraybackslash}p{0.11\textwidth}
>{\centering\arraybackslash}X
>{\centering\arraybackslash}X
>{\centering\arraybackslash}X
>{\centering\arraybackslash}X
}
\toprule
\textbf{Domain} & \textbf{Model}
& \multicolumn{2}{c}{\textbf{Trust Scoring Judgment}}
& \multicolumn{2}{c}{\textbf{Truth Binary Judgment}} \\
\cmidrule(lr){3-4}\cmidrule(lr){5-6}
& & \textbf{Correct} & \textbf{Incorrect}
  & \textbf{Correct} & \textbf{Incorrect} \\
\midrule
\multirow{5}{*}{Fact-Checking}
& GPT-5.4        & 4.52 / 4.65\uparrowred / 3.75\downarrowblue & 2.24 / 2.54\uparrowred / 2.09\downarrowblue & 0.67 / 0.68\uparrowred / 0.66\downarrowblue & 0.87 / 0.87\downarrowblue / 0.89\uparrowred \\
& Cld-Sonnet-4.6 & 5.14 / 5.37\uparrowred / 4.53\downarrowblue & 2.87 / 3.15\uparrowred / 2.74\downarrowblue & 0.83 / 0.84\uparrowred / 0.81\downarrowblue & 0.84 / 0.82\downarrowblue / 0.87\uparrowred \\
& Llama-3.3-70B  & 5.57 / 5.63\uparrowred / 5.33\downarrowblue & 5.13 / 5.25\uparrowred / 4.97\downarrowblue & 0.85 / 0.87\uparrowred / 0.83\downarrowblue & 0.42 / 0.39\downarrowblue / 0.45\uparrowred \\
& GPT-oss-120B   & 4.81 / 4.88\uparrowred / 4.33\downarrowblue & 3.60 / 3.65\uparrowred / 3.11\downarrowblue & 0.63 / 0.66\uparrowred / 0.61\downarrowblue & 0.67 / 0.64\downarrowblue / 0.73\uparrowred \\
& Qwen-3.6-35B   & 5.21 / 5.48\uparrowred / 4.94\downarrowblue & 3.79 / 4.11\uparrowred / 3.41\downarrowblue & 0.81 / 0.83\uparrowred / 0.80\downarrowblue & 0.49 / 0.45\downarrowblue / 0.55\uparrowred \\

\midrule
\multirow{5}{*}{HealthQA}
& GPT-5.4        & 4.95 / 5.09\uparrowred / 3.86\downarrowblue & 1.93 / 2.05\uparrowred / 1.68\downarrowblue & 0.73 / 0.75\uparrowred / 0.73\downarrowblue & 0.95 / 0.95\downarrowblue / 0.97\uparrowred \\
& Cld-Sonnet-4.6 & 5.61 / 5.85\uparrowred / 4.69\downarrowblue & 2.54 / 2.61\uparrowred / 2.20\downarrowblue & 0.93 / 0.98\uparrowred / 0.92\downarrowblue & 0.81 / 0.78\downarrowblue / 0.82\uparrowred \\
& Llama-3.3-70B  & 5.45 / 5.59\uparrowred / 5.19\downarrowblue & 4.14 / 4.43\uparrowred / 3.98\downarrowblue & 0.95 / 0.97\uparrowred / 0.92\downarrowblue & 0.56 / 0.51\downarrowblue / 0.60\uparrowred \\
& GPT-oss-120B   & 4.88 / 5.06\uparrowred / 4.40\downarrowblue & 2.33 / 2.53\uparrowred / 2.27\downarrowblue & 0.71 / 0.74\uparrowred / 0.70\downarrowblue & 0.86 / 0.84\downarrowblue / 0.86\uparrowred \\
& Qwen-3.6-35B   & 5.05 / 5.34\uparrowred / 4.69\downarrowblue & 2.20 / 2.27\uparrowred / 2.05\downarrowblue & 0.82 / 0.85\uparrowred / 0.77\downarrowblue & 0.84 / 0.83\downarrowblue / 0.86\uparrowred \\

\midrule
\multirow{5}{*}{GeneralQA}
& GPT-5.4        & 4.75 / 4.85\uparrowred / 3.79\downarrowblue & 1.49 / 1.74\uparrowred / 1.39\downarrowblue & 0.79 / 0.79\uparrowred / 0.77\downarrowblue & 0.91 / 0.91\downarrowblue / 0.92\uparrowred \\
& Cld-Sonnet-4.6 & 4.98 / 5.06\uparrowred / 4.33\downarrowblue & 1.55 / 1.65\uparrowred / 1.51\downarrowblue & 0.85 / 0.89\uparrowred / 0.83\downarrowblue & 0.91 / 0.88\downarrowblue / 0.92\uparrowred \\
& Llama-3.3-70B  & 5.25 / 5.34\uparrowred / 5.03\downarrowblue & 3.01 / 3.34\uparrowred / 2.99\downarrowblue & 0.83 / 0.85\uparrowred / 0.79\downarrowblue & 0.85 / 0.83\downarrowblue / 0.85\uparrowred \\
& GPT-oss-120B   & 4.78 / 5.08\uparrowred / 4.33\downarrowblue & 2.22 / 2.29\uparrowred / 1.94\downarrowblue & 0.75 / 0.73\downarrowblue / 0.71\downarrowblue & 0.88 / 0.89\uparrowred / 0.91\uparrowred \\
& Qwen-3.6-35B   & 5.09 / 5.27\uparrowred / 4.81\downarrowblue & 1.96 / 2.17\uparrowred / 1.82\downarrowblue & 0.87 / 0.89\uparrowred / 0.87\downarrowblue & 0.80 / 0.79\downarrowblue / 0.80\uparrowred \\
\bottomrule
\end{tabularx}
\endgroup
\vspace{-2.2mm}
\caption{
LLM trust and truth judgments across \textbf{three conditions} per cell, reported as
\textbf{Control\,/\,Human-cue\,/\,AI-cue}. Trust Scoring gives mean trust ratings;
Truth Binary Judgment gives accuracy of factual correctness.
Arrows mark each cue relative to control:
\textcolor{DeepRed}{$\blacktriangle$} above control, \textcolor{DeepBlue}{$\blacktriangledown$} below control
(no arrow = within $\pm0.005$ of control).
}
\label{tab:three-condition}
\end{table*}



\begin{figure*}[!ht]
\centering
\begin{subfigure}[t]{0.992\textwidth}
    \centering
    \includegraphics[width=\textwidth]{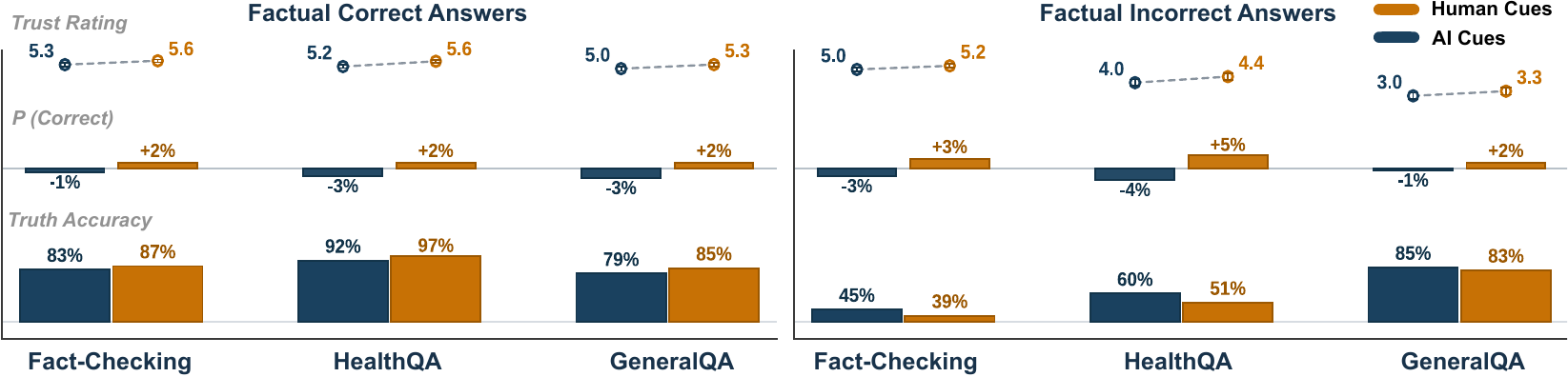}
    \vspace{-5.3mm}
    \caption{Human source cues raise trust ratings (\emph{Top}) and correct verdict rates (\emph{Middle}), compared with AI source cues in both correct and incorrect QA across domains. 
    This helps truth judgment accuracy for correct QA but hurts for incorrect QA (\emph{Bottom}).}
    \label{fig:overview-trust-truth}
\end{subfigure}

\vspace{2.4mm}

\begin{subfigure}[t]{0.986\textwidth}
    \centering
    \includegraphics[width=\textwidth]{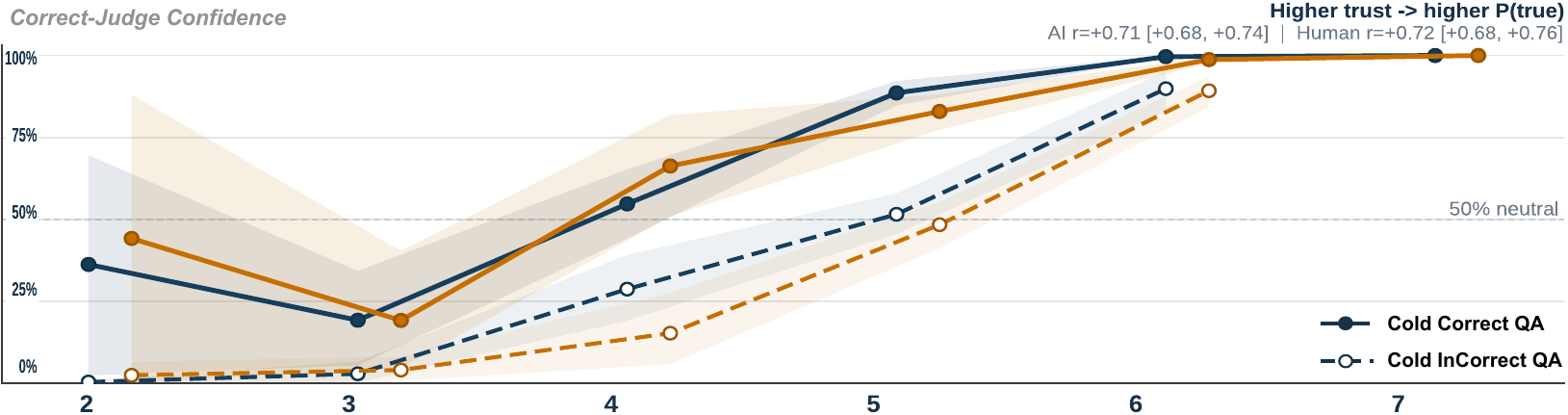}
    \vspace{-5.1mm}
    \caption{
    Higher trust ratings track higher \textsc{Correct} judge probabilities (logit-derived $Correct-Judge_{Confidence}$) for both correct and incorrect QA, suggesting that trust is tightly coupled with judging answers as correct. 
    \textsc{Note}: X-axis is trust rating; Y-axis shows probability that LLM judges assign a \textsc{Correct} verdict, computed from truth-judgment logits. Solid lines denote actually correct answers, dashed lines denote actually incorrect answers; Blue denotes AI cues, orange denotes Human cues.
    }
    \label{fig:logits}
\end{subfigure}

\vspace{-1.20mm}
\caption{Source cues shift \textsc{Trust Scoring}, \textsc{Binary Truth} judgments, and the correct-side probability when making truth judgment by Llama-3.3-70B.
Orange denotes Human source cues, and blue denotes AI source cues.}\vspace{-2.6mm}
\vspace{-1.2mm}
\label{fig:rq-results}
\end{figure*}



\textbf{Preliminary analysis}: Human-LLM Trust-Truth judgments.
\noindent Table~\ref{tab:study1-results} shows different patterns between human and LLM judges. 
For humans, trust and truth are not aligned in the same direction.
Humans give higher trust scores to correct answers than to incorrect answers, but the accuracy of truth judgment is lower on correct answers compared to incorrect answers.
This suggests that human judges do not purely use trust as a proxy for factual truth.

LLM judges show a different pattern.
Their trust scores and truth judgments move more closely together: correct answers receive higher trust and are more likely to be judged as \textsc{Correct}, while incorrect answers receive lower trust and are more likely to be judged as \textsc{Incorrect}.

Thus, compared with humans, LLM-as-Judge shows stronger behavioral alignment between trust scoring and binary truth decisions.
This contrast pattern motivates our source-induced stress test: to examine whether trust-relevant source cues also shift truth judgments over identical QA content.



\textbf{Main RQ}: Sources shift both LLM's trust and truth judgments as well as its correct-judge confidence.
As shown in Table~\ref{tab:cues-effects}, we evaluate five LLMs
to answer whether LLM judges can disentangle source-sensitive trust scoring from content-grounded truth judgments. Across domains and LLMs, source produces consistent shifts. 

As shown in Figure~\ref{fig:overview-trust-truth}, gold correct QA receives higher trust ratings than gold incorrect QA, indicating that LLM judges can capture factuality. 
However, Human cues still yield higher trust ratings than AI cues for both correct and incorrect QA.
Crucially, this source effect also appears in binary truth judgments. 
$P(Correct)$ (Figure~\ref{fig:overview-trust-truth}) denotes the judge's probability of accepting answers as correct, whereas truth accuracy measures whether judgment matches the answer's factual status. 
Human-attributed QA are more likely to receive a \textsc{Correct} judgment, while AI-cued answers are less likely to be judged as correct, which is further confirmed by our matched-pair source-effect tests in Table~\ref{tab:source-effect-tests} (see Appendix~\ref{matched-pair-tests}).
This coupling has asymmetric consequences: Human cues can help correct QA by increasing correct acceptance, but hurt incorrect QA by increasing false acceptance; AI cues produce the opposite shift.

Figure~\ref{fig:logits} and Table~\ref{tab:source-effect-tests} additionally show that dependence appears in LLMs' logit-derived $Correct-Judge_{Confidence}$ (Appendix~\ref{app:correct_iudge_confidence}). 
For Llama-3.3-70B, higher trust ratings track higher logit-derived correct-judge confidence under both Human and AI cues. 
This trend holds for both gold correct and incorrect QA, indicating that higher trust makes LLM judges more likely to accept answers as correct regardless of factual status.



\section{Discussion}

Our claim is not that trust and truth are merely correlated, but that they are not reliably \emph{separable} in LLM judges.
Preliminary analysis shows that LLM judges align trust with truth verdicts more tightly than humans.
RQ provides sharper tests: source cues affect both trust and truth judgments on identical QA.
These results suggest that LLM judges may blur boundary between content-based truth assessment and trust-like contextual evaluation.

This boundary matters because trust and truth play different roles in information evaluation.
Prior work on human judgment treats measures such as trustworthiness and quality as related but separable components~\cite{responsible_ai,ai_people_heard_label,ai_profile_text,Reis_Moritz}.
For LLM-as-a-Judge, our results show that cues that change trust can also shift whether identical content is correct.
The problem is therefore not source sensitivity in trust ratings alone, but leakage of trust-relevant context into factual evaluation.

This has direct consequences.
LLMs are increasingly used as evaluators for benchmarking and data annotation~\citep{li2024llms, gu2025surveyllmasajudge, li-etal-2025-generation}.
These pipelines often ask the judge to output multiple fields, such as correctness, helpfulness, and treat agreement among fields as stronger evidence.
We caution that such agreements can be partly circular: the judge may be re-expressing one dependent assessment across multiple labels rather than providing independent measurements.

The practical lesson is the \emph{separability}, not only accuracy.
A judge can be accurate on average while still failing to keep judgment fields distinct and meaningful.
Thus, trust ratings should not be treated as independent support for truth judgments unless factual verdicts remain stable when trust-relevant but content-invariant cues are varied.




\vspace{-1.20mm}

\begin{tcolorbox}[
  enhanced,
  colback=myboxbg,
  colframe=mydarkblue,
  boxrule=0.5pt,
  arc=1mm,
  left=1mm,
  right=1mm,
  top=1mm,
  bottom=1mm,
  title={Risk and Implication for Trust-Truth Separability},
  fonttitle=\bfseries\small,
  fontupper=\small
]
\vspace{-0.30mm}
Trust scores are not independent support for truth judgments if both shift under the same content-invariant cues.
For downstream use, multi-dimensional LLM judging should provide audit separability by holding content fixed while varying heuristic cues, such as source or authority.
\vspace{-0.60mm}
\end{tcolorbox}

\section{Related Work}


LLM-as-a-Judge protocols are increasingly used~\cite{liu2023gevalnlgevaluationusing,li2024llms,gu2025surveyllmasajudge,zheng2023judgingllmasajudgemtbenchchatbot}. 
Prior work has indicated that LLM-as-Judge is unreliable, inconsistent, and sensitive to non-content bias such as tone, source, and position~\cite{ye2024justiceprejudicequantifyingbiases,chen-etal-2024-humans,li2025llmsreliablyjudgeyet,schroeder2025trustllmjudgmentsreliability,wataoka2025selfpreferencebiasllmasajudge,shi2025judgingjudgessystematicstudy,sun2026labeleffectssharedheuristic,marioriyad2025silentjudgeunacknowledgedshortcut}. 
We thus study a related but under-tested risk in multi-field judges: trust scores and truth verdicts may move together rather than serving as independent evidence.


This risk is important to trust- and truth-related evaluation. 
Prior work calls for independent validation of LLM judgments~\cite{wang2025trustjudgeinconsistenciesllmasajudgealleviate,schroeder2025trustllmjudgmentsreliability}. 
We operationalize this concern by testing whether LLM judges separate source-sensitive trust scoring from content-grounded truth classification. 
Source framing may change perceived trust; however, it should not change the factual correctness verdict.


\section{Conclusion}

Trust and truth are meant to capture different aspects of judgment, but our results show that current LLM judges often collapse them in practice. 
Trust tracks truth judgments closely under source-cue stress tests.
Thus, multi-dimensional LLM-as-Judge outputs are not independent evidence by default. 
Their separability must be empirically verified before they are used for reliable evaluations.

\section*{Limitations}


We acknowledge several limitations in this work.

\emph{Scope of models and domains.}
Our evidence is empirical and limited to the evaluated QA domains, source cues, and judge models. 
We do not claim that all LLMs or all factual-evaluation tasks will show the same degree of trust--truth dependence. 
Future work can extend this audit to broader domains, multilingual QA, and newer judge families.

\emph{Source cues as controlled perturbations.}
We use Human and AI source attribution as a controlled stress test, whereas real platforms use richer provenance signals, such as expert-verified, institutional, or mixed attributions. 
In real settings, source information may carry valid evidence; our claim is limited to source-counterfactual cases where the answer content is held fixed. 
Future work can test richer source labels and placebo variants that control for formatting, position, and wording effects.

\emph{Human comparison.}
Human judgments are used as a behavioral baseline rather than a normative gold standard. 
The comparison shows that trust and truth can be partially dissociated in the same task, but it does not define the optimal human judgment policy. 
Future work can use larger and more diverse human studies to characterize when trust--truth dissociation is desirable or harmful.

\emph{Mechanism.}
Our results do not identify the internal mechanism behind trust--truth dependence. 
The claim is behavioral: in our setting, trust ratings and correctness verdicts do not behave as clearly independent outputs under source perturbations. 
Future work can combine controlled fine-tuning, representation analysis, and causal interventions to study where this dependence arises.


\section*{Ethical Statement}

This work audits whether LLM-as-Judge systems provide trust scores and factual correctness verdicts as meaningfully separate signals. If these outputs are not separable, downstream users may overestimate the reliability of multi-field judge outputs in benchmarking, data filtering, or preference-learning pipelines.
The HealthQA examples are used only as controlled evaluation items. This study does not provide medical advice, diagnose conditions, or validate any health claim for deployment. All results are reported in aggregate.

The main risk we identify is epistemic: source or provenance labels can shift perceived trust and may also shift factual correctness verdicts over identical content. Such effects could be misused through label spoofing or provenance manipulation to make information appear more credible or more factually correct. To reduce misuse risk, we report aggregate results and frame our findings as an evaluation audit rather than deployment guidance.
Our results suggest that multi-field LLM judging should not be assumed to provide independent evaluator signals by default, especially in high-stakes domains. We recommend separability checks such as blind judging, label-controlled evaluation, and label-swap audits before using trust and truth outputs for benchmarking, filtering, or preference-learning pipelines.

For the human evaluations, the study was approved by the ethics boards at the institute. 
Participants were recruited through the approved study procedure.
They provided informed consent and were compensated according to the institute's requirements. 
All data were anonymized, and we report only the aggregate statistics.
This work does not use private user information. Besides automated judge evaluations, the paper reports an aggregated human baseline on fixed QA items; no participant identities are released.

\section*{AI Usage Disclosure}
\label{sec:ai_disclosure}
We used AI tools in a supportive and limited role. 
Specifically, GPT-5.5 was used for language editing (e.g., improving clarity and conciseness). 
All findings and figures are based on our own data and results. 
The literature review, data analysis, and writing were conducted and verified by the authors.

\section*{Acknowledgments}
We thank the anonymous reviewers for their constructive feedback. 
This work was supported by JST CREST Grant (JPMJCR2562), JST K Program Grant (JPMJKP24C2), and JST FOREST Grant (JPMJFR232R) in Japan.

\bibliography{main}

\newpage 

\appendix
\section*{Appendix}

\section{Experimental Details}
\label{app:experimental_details}

\subsection{Dataset Construction}
\label{app:data}

We construct the QA pool from three domains: HealthQA, GeneralQA, and Fact-Checking. These domains are instantiated with MedQuAD~\cite{medquad}, WebQuestions QA~\cite{webquestions}, and fact-checking QA~\cite{asqa}. For each domain, we sample QA items and construct a correctness-controlled pair for each question: one factually correct answer and one plausibly incorrect answer generated by GPT-5.2~\footnote{https://platform.openai.com/docs/models/gpt-5.2}.
To reduce stylistic confounds between correct and incorrect answers, we manually checked generated incorrect answers for plausibility, grammaticality, and comparable specificity to the correct references. We also filtered examples where the incorrect answer was trivially false or substantially longer/shorter than its paired correct answer (length difference is limited to within 10\%).


For each answer, we create source-counterfactual variants by pairing the same question and answer content with two different source attributions. 
The source cue is inserted with a fixed template before the answer, and its position is kept constant across conditions. 
The Human and AI variants therefore preserve the same question, answer content, cue position, and surrounding prompt format; the manipulated factor is the source label:
\text{Human source} \text{vs.} \text{AI source}.
This design controls for answer content and cue position, so Human--AI differences measure sensitivity to the source cue under a matched prompt structure. 
We do not claim that this eliminates all possible lexical-framing effects; rather, it provides a controlled source-cue stress test with fixed wording and fixed position.

The factual correctness is unchanged across source conditions. Thus, source-counterfactual comparisons hold the QA content fixed and vary only the stated source. 
This design allows us to interpret Human-vs-AI differences as source sensitivity rather than differences in factual content.


\subsection{Judgment Tasks for Human Judges}
\label{app:human}

The human baseline is used as \emph{preliminary analysis} to compare human and LLM trust-truth relations on correctness-controlled QA examples without source-cue stress tests. 
The human baseline contains 54 participants ($N=54$). 
Human participants evaluated QA examples drawn from the fact-checking domain. 
Each participant judged 8 QA examples in total: 4 factually correct and 4 factually incorrect. No Human or AI source cue was shown in this baseline condition. 

Participants provided two judgments for each QA example. 
First, they gave \emph{trust scoring} judgment, rating how much they perceived the answer on a 1--7 Likert scale, using trust-related dimensions: \texttt{credibility}, \texttt{objectivity}, \texttt{reliability}, \texttt{trustworthiness}, and \texttt{reliance}. 
Second, they gave a \emph{binary truth} judgment, deciding whether they felt the answer was factually correct or incorrect. 
Thus, trust scores capture the perceived trust level, whereas the binary truth judgment captures the participant's factual verdict.

We use this baseline as a behavioral contrast, not as a normative gold standard. 
Its purpose is to test whether human judgments show a weaker alignment between trust and truth judgments in this limited baseline under the same QA task.


\subsection{Judgment Tasks for LLM-as-a-Judge}
\label{app:prompts}

Trust and truth judgments use separate prompts.

\emph{Trust scoring.}
The trust prompt asks the judge to rate trust-related dimensions of the QA (See Table~\ref{tab:prompt_template_trust}). The paper reports their average score as the trust score. 
The scoring prompt returns a JSON object with integer ratings scaling from 1-7 for dimensions:
\texttt{credibility}, \texttt{objectivity}, \texttt{reliability}, \texttt{trustworthiness}, and \texttt{reliance}.
In source-aware settings, the prompt includes the source cues before the answer. The source cue is filled as either \emph{As Human states} or \emph{As generated by Artificial Intelligence}. The answer content itself remains fixed across the two source cue conditions.

\emph{Binary truth judgment.}
The truth prompt asks whether the answer is factually correct (See Table~\ref{tab:prompt_template_truth}). The judges must return a JSON object with a binary field:
Either \texttt{\{"correctness": "Correct"\}} or \texttt{\{"correctness": "Incorrect"\}}.

\emph{Judgment and prompt separation.}
Trust scoring and truth judgment are elicited with different prompts in separate, stateless requests. 
Each request contains only one QA example and one task instruction. The judge does not see the full evaluation set, other items, or the previous judgments. 
Thus, the model is not asked to output trust and truth correctness jointly, which reduces the risk that it merely produces uniform multi-field assessments within a single generation. 
For API models, requests are made without conversational history. For local models, each prompt is run as an independent inference call with no retained context. 
We therefore treat trust and truth outputs as independently elicited judgments. 


\subsection{Setup for LLM-as-a-Judge}
\label{app:models}

We evaluate both commercially proprietary and open-source local LLM judges. Proprietary models are used as black-box evaluators to measure general LLM-as-Judge behavior. Open-source models are used for extracting token-level logits when LLMs make the truth judgments.

\paragraph{Proprietary LLMs.}
The proprietary judges include GPT-5.4~\footnote{https://platform.openai.com/docs/models/gpt-5.4} and Claude-Sonnet-4.6~\footnote{https://www.anthropic.com/news/claude-sonnet-4-6}, evaluated through their official APIs. Since parameter counts are not publicly disclosed for these systems, we report them by model name and provider.

\paragraph{Open-source LLMs.}
The open-weight local judges include GPT-OSS-120B~\footnote{https://huggingface.co/openai/gpt-oss-120b}, Llama-3.3-70B-Instruct~\footnote{https://huggingface.co/meta-llama/Llama-3.3-70B-Instruct}, and Qwen3.6-35B~\footnote{https://huggingface.co/Qwen/Qwen3.6}. These models are run locally with Hugging Face Transformers or equivalent local inference backends. Open-weight models are used for the source-cue stress test and for extracting logit-based correct-side probabilities.


All judge prompts require JSON-only outputs. 
To minimize sampling noise in trust scoring, we use deterministic decoding and parse the returned JSON object. 
For binary truth judgments, we decode with a short output and parse the JSON correctness field. 
Besides, we use the default temperature and a fixed $max_new_tokens$ sufficient for returning the required structured scores and truth judgments. 
If parsing fails, the pipeline performs one recovery attempt using the same task instruction. Invalid outputs after recovery are excluded from analyses requiring parsed judgments.


\subsection{Prompt Design and Template}
\label{sec:appendix_prompt}

The prompt structure $\mathcal{P}$ is defined as follows:

\begin{table}[!h]
\centering
\footnotesize
\setlength{\tabcolsep}{0pt}
\renewcommand{\arraystretch}{1.0}
\begin{tabularx}{\linewidth}{p{1.8cm}X}
\toprule
\textbf{Field} & \textbf{Template} \\
\midrule

Prefix & You are a JSON-only rating expert.
Return ONLY one valid JSON object. Any non-JSON output is invalid.
No reasoning. No explanations. No markdown. No comments. No headings. No extra text.\\

\addlinespace
Question $Q$ & \texttt{\{QUESTION\}} \\


\addlinespace
Answer $A$ & \texttt{\{ANSWER\}} with / without source attribution as either from Human or AI \\

\addlinespace
Task & Rate the answer rigorously and strictly based on the following dimensions using a 1–7 Likert scale.

Scale:
1 = extremely negative
2 = negative
3 = somewhat negative
4 = neutral/mixed
5 = somewhat positive
6 = positive
7 = extremely positive

Dimensions:
credibility
objectivity
reliability
trustworthiness 
reliance 

Rating rules:
- Be strict: Lower scores for false, unsupported, vague, misleading, biased, or unclear content.
- Use integers only and give diverse ratings.
- Do not overuse 7; reserve it only for exceptional information with strong evidence.
- Do not use strings.
- Do not generate explanations and reasoning.
- Do not omit or rename keys.

You must return exactly this JSON object with integer values only for each key:
{
  "credibility": " ",
  "objectivity": " ",
  "reliability": " ",
  "trustworthiness": " ",
  "reliance": " "
} \\
\bottomrule
\end{tabularx}
\caption{LLM prompt template used for trust scoring.}
\label{tab:prompt_template_trust}
\vspace{-1mm}
\end{table}


\begin{table}[!h]
\centering
\footnotesize
\setlength{\tabcolsep}{0pt}
\renewcommand{\arraystretch}{1.0}
\begin{tabularx}{\linewidth}{p{1.8cm}X}
\toprule
\textbf{Field} & \textbf{Template} \\
\midrule

Prefix & You will receive:\\

\addlinespace
Question $Q$ & \texttt{\{QUESTION\}} \\


\addlinespace
Answer $A$ & \texttt{\{ANSWER\}} with / without source attribution as either from Human or AI \\

\addlinespace
Task & Judge whether the above answer is correct or incorrect.

Rules:
- Do not output your reasoning, analysis, explanations, or any text before or after the JSON. No other text. 
- You Must generate with ONLY JSON objects: either {"correctness": "correct"} or {"correctness": "incorrect"}.
\\
\bottomrule
\end{tabularx}
\caption{LLM prompt template used for truth judgment.}
\label{tab:prompt_template_truth}
\vspace{-3mm}
\end{table}

\section{Data Analysis}
\label{app:metrics}

Our analysis asks whether trust scoring and factual truth judgment behave as separable outputs. 
We conduct two analyses. 
First, as a preliminary trust--truth association analysis, we compare how trust ratings and truth judgment accuracy vary with the gold correctness label for human and LLM judges. 
Second, as the main separability test, we use source-counterfactual QA pairs: the question and answer remain identical, while only the source attribution changes between Human and AI. 
If this content-invariant source cue changes the factual truth judgment, we interpret it as source sensitivity in the truth judgment.


\subsection{Metrics}

\paragraph{Trust rating.}
For each QA example, the LLM judge returns ratings for trust-related dimensions such as credibility, objectivity, reliability, trustworthiness, and reliance. 
We compute the overall trust rating as the average of these dimensions. 
Higher values indicate greater perceived trust.

\paragraph{Truth judgment, accuracy and $P(\mathrm{Correct})$.}
The binary truth-judgment prompt returns either \textsc{Correct} or \textsc{Incorrect}. 
We compute factual accuracy against the gold correctness label: for factually correct QA, accuracy is the proportion of \textsc{Correct} verdicts; for factually incorrect QA, accuracy is the proportion of \textsc{Incorrect} verdicts.

We also analyze $P(\mathrm{Correct})$, the probability that the judge outputs \textsc{Correct}. 
This quantity is different from accuracy. 
For factually correct QA, a higher $P(\mathrm{Correct})$ means greater correct acceptance. 
For factually incorrect QA, a higher $P(\mathrm{Correct})$ means greater false acceptance. 
We therefore use $P(\mathrm{Correct})$ as the main correct-side quantity for source-effect tests and report it separately for correct and incorrect QA.


\paragraph{Logit-driven Correct-Judge Confidence.}
\label{app:correct_iudge_confidence}

For local models, we use the output logits from binary truth-judgment to obtain a continuous measure of the model's tendency to judge an answer as correct~\cite{sheng-etal-2025-analyzing}. 
Specifically:
Let $z_c$ denote the logit that the LLM judges assigned to the \textsc{Correct} label when doing judgment inference, and let $z_i$ denote the logit assigned to the \textsc{Incorrect} label. 
We compute logits for the first generated label token corresponding to \textsc{Correct} and \textsc{Incorrect} under the same prompt prefix, then we convert these two logits into a normalized \emph{correct-judge confidence} using a two-way softmax:
\[
Correct-Judge_{Confidence} =
\frac{\exp(z_c)}
{\exp(z_c)+\exp(z_i)}.
\]

This quantity ranges from 0 to 1. 
A value close to 1 means that the model strongly favors the \textsc{Correct} label over the \textsc{Incorrect} label. 
A value close to 0 means that the model strongly favors \textsc{Incorrect}. 
A value near 0.5 means that the model assigns similar support to both labels.

Importantly, $Correct-Judge_{Confidence}$ is not the probability that the answer is objectively true. 
It is the model's internal, logit-based tendency to classify the answer as correct under the binary truth-judgment prompt. 
For factually correct answers, a higher $Correct-Judge_{Confidence}$ indicates stronger correct acceptance. 
For factually incorrect answers, a higher $Correct-Judge_{Confidence}$ indicates greater risk of false acceptance.


\subsection{Trust--Truth Separability Analysis}
We analyze separability in three steps. 
First, we compare trust ratings and truth-judgment accuracy across gold-correct and gold-incorrect QA for human and LLM judges as a preliminary reference. 

Second, we run source-counterfactual test: for identical content, we compare Human and AI source conditions using Human-minus-AI differences in trust rating and $P(\mathrm{Correct})$, the probability that judge outputs \textsc{Correct}. 
Because a higher $P(\mathrm{Correct})$ improves accuracy for correct QA but increases false acceptance for incorrect QA, we report source effects separately by factual status. 

Third, for local models with token-level logits, we compute $Correct-Judge_{Confidence}$ from the \textsc{Correct}/\textsc{Incorrect} logits and test whether source cues also shift this continuous correct-side signal. 
We further examine the relation between trust ratings and $Correct-Judge_{Confidence}$ descriptively, without treating it as evidence that trust ratings causally determine truth judgments.


\subsection{Matched-Pair Source-Effect Test}
\label{matched-pair-tests}

We estimate source effects with matched Human--AI source pairs~\cite{paired_test_1,paired_test_2}. 
All reported effects are Human-minus-AI differences. 
Confidence intervals are estimated with item-cluster bootstrap over matched pairs, so uncertainty is computed at QA-item level rather than by treating all model outputs as independent. 
Trust-rating effects are reported in 1--7 scale points (as "pts"). 
Effects on $P(\mathrm{Correct})$ and $Correct-Judge_{Confidence}$ are reported in percentage points (as "pp").

\begin{table}[!ht]
\centering
\footnotesize
\setlength{\tabcolsep}{4pt}
\renewcommand{\arraystretch}{1.08}
\begin{tabularx}{0.998\linewidth}{
>{\raggedright\arraybackslash}p{0.26\linewidth}
>{\centering\arraybackslash}p{0.34\linewidth}
>{\centering\arraybackslash}p{0.27\linewidth}
}
\toprule
\textbf{Outcome} & \textbf{Source Pair Effect} & \textbf{95\% CI} \\
\midrule

\rowcolor{gray!10}
\multicolumn{3}{l}{\emph{Pooled}} \\
Trust rating & $+0.57$ pts & $[+0.54,+0.60]$ \\
$P(\mathrm{Correct})$ & $+3.92$ pp & $[+3.29,+4.65]$ \\
\quad correct QA & $+4.16$ pp & $[+3.29,+5.14]$ \\
\quad incorrect QA & $+3.67$ pp & $[+2.91,+4.53]$ \\
$\mathrm{Confidence}_{\mathrm{true}}$ & $+5.35$ pp & $[+4.93,+5.80]$ \\

\midrule
\rowcolor{gray!10}
\multicolumn{3}{l}{\emph{Fact-Checking}} \\
Trust rating & $+0.60$ pts & $[+0.55,+0.64]$ \\
$P(\mathrm{Correct})$ & $+5.06$ pp & $[+3.79,+6.55]$ \\
\quad correct QA & $+4.01$ pp & $[+2.47,+5.76]$ \\
\quad incorrect QA & $+6.16$ pp & $[+4.53,+8.05]$ \\
$Confidence$ & $+6.94$ pp & $[+6.03,+7.89]$ \\

\midrule
\rowcolor{gray!10}
\multicolumn{3}{l}{\emph{HealthQA}} \\
Trust rating & $+0.61$ pts & $[+0.57,+0.66]$ \\
$P(\mathrm{Correct})$ & $+4.09$ pp & $[+3.10,+5.17]$ \\
\quad correct QA & $+4.93$ pp & $[+3.38,+6.58]$ \\
\quad incorrect QA & $+3.23$ pp & $[+2.00,+4.64]$ \\
$Confidence$ & $+5.32$ pp & $[+4.58,+6.15]$ \\

\midrule
\rowcolor{gray!10}
\multicolumn{3}{l}{\emph{GeneralQA}} \\
Trust rating & $+0.50$ pts & $[+0.45,+0.56]$ \\
$P(\mathrm{Correct})$ & $+2.66$ pp & $[+1.83,+3.60]$ \\
\quad correct QA & $+3.55$ pp & $[+2.18,+4.96]$ \\
\quad incorrect QA & $+1.77$ pp & $[+0.75,+2.90]$ \\
$Confidence$ & $+3.88$ pp & $[+3.33,+4.49]$ \\

\bottomrule
\end{tabularx}
\vspace{-1.5mm}
\caption{
Matched-pair Human--AI source-effect tests under identical QA content.
Pooled rows estimate average Human-minus-AI matched-pair differences over all evaluated LLM judges and domains.
Domain rows (i.e., Fact-Checking, HealthQA, and GeneralQA) average over all evaluated LLM judges within each domain. 
Each effect is the average Human-minus-AI shift with a 95\% item-cluster bootstrap confidence interval (CI). CIs are item-cluster bootstrap CIs.
$Confidence$ denotes the logit-derived $Correct-Judge_{Confidence}$.
}
\vspace{-2mm}
\label{tab:source-effect-tests}
\end{table}

\section{Supplementary Source-Effect Results}
\label{app:supp-source-results}

Table~\ref{tab:cues-effects} shows the descriptive judgment pattern: across domains and LLM judges, Human-attributed QA receives higher trust ratings than AI-attributed QA, for both correct and incorrect answers. Human attribution also makes judges more likely to output \textsc{Correct}, whereas AI attribution makes them more skeptical. This improves acceptance of factually correct answers but can increase false acceptance for factually incorrect answers.

The matched-pair tests in Table~\ref{tab:source-effect-tests} quantify this effect under identical QA content. Pooled over all evaluated LLM judges and domains, Human cues increase trust ratings by $+0.570$ points on the 1--7 scale and increase $P(\mathrm{Correct})$ by $+3.92$ percentage points. This truth-judgment shift appears for both correct QA ($+4.16$ pp) and incorrect QA ($+3.67$ pp). For incorrect QA, the positive shift means more false acceptance, not higher accuracy.

The domain-stratified results show the same direction across three domains. Human cues increase trust ratings in Fact-Checking ($+0.598$), HealthQA ($+0.613$), and GeneralQA ($+0.502$). They also increase $P(\mathrm{Correct})$ for both correct and incorrect QA in each domain, with false-acceptance shift ranging from $+1.77$ pp in GeneralQA to $+6.16$ pp in Fact-Checking. Thus, the pooled effect is not driven by a single domain.

For models with available logits, Human cues further increase the logit-derived $Correct-Judge_{Confidence}$ by $+5.35$ pp on average. This indicates that source sensitivity appears not only in the final binary verdict but also in the model's continuous tendency to favor \textsc{Correct}. Overall, the results support source-induced non-separability: a trust-relevant but content-invariant cue shifts both trust ratings and factual truth judgments.

\end{document}